\pdfoutput=1
\documentclass[11pt]{article}

\usepackage{ACL2023}
\usepackage{times}
\usepackage{latexsym}

\usepackage[T1]{fontenc}
\usepackage[utf8]{inputenc}

\usepackage{microtype}

\usepackage{inconsolata}

\usepackage[utf8]{inputenc} % allow utf-8 input
\usepackage[T1]{fontenc}    % use 8-bit T1 fonts
\usepackage{hyperref}       % hyperlinks
\usepackage{url}            % simple URL typesetting
\usepackage{booktabs}       % professional-quality tables
\usepackage{amsfonts}       % blackboard math symbols
\usepackage{nicefrac}       % compact symbols for 1/2, etc.
\usepackage{microtype}      % microtypography
\usepackage{xcolor}         % colors
\usepackage{adjustbox}

\usepackage{enumitem}
\usepackage{algorithm}
\usepackage{algpseudocode}
\usepackage{bbm}

\usepackage{xspace}

\makeatletter
\DeclareRobustCommand\onedot{\futurelet\@let@token\@onedot}
\def\@onedot{\ifx\@let@token.\else.\null\fi\xspace}

\def\eg{\emph{e.g}\onedot} 
\def\ie{\emph{i.e}\onedot} 

\newcommand{\Ni}{({\em i})~}
\newcommand{\Nii}{({\em ii})~}
\newcommand{\Niii}{({\em iii})~}

\usepackage{amsmath,amsfonts,bm}
\usepackage[nameinlink]{cleveref}
\crefformat{section}{\S#2#1#3} % see manual of cleveref, section 8.2.1
\crefname{algorithm}{Alg.}{Algs.}
\crefformat{subsection}{\S#2#1#3}
\Crefname{equation}{Eq.}{Eqs.}
\Crefname{figure}{Fig.}{Figs.}

\def\eqref#1{equation~\ref{#1}}
\def\1{\bm{1}}

\DeclareMathAlphabet{\mathsfit}{\encodingdefault}{\sfdefault}{m}{sl}
\SetMathAlphabet{\mathsfit}{bold}{\encodingdefault}{\sfdefault}{bx}{n}

\usepackage{hyperref}
\usepackage{url}

\usepackage{multirow}
\usepackage{makecell}
\usepackage{graphicx}
\usepackage{tablefootnote}
\usepackage{adjustbox}
\usepackage{subfiles}
\usepackage{booktabs}
\usepackage{cleveref}
\usepackage{wrapfig}
\usepackage{todonotes}
\usepackage{enumitem}

\usepackage{subcaption}
\usepackage{latexsym}
\usepackage{color, colortbl}
\usepackage{amsmath}
\usepackage{amssymb}

\crefformat{section}{\S#2#1#3}
\crefformat{subsection}{\S#2#1#3}

\crefformat{subsubsection}{\S#2#1#3}
\title{Verify-and-Edit: A Knowledge-Enhanced Chain-of-Thought Framework}

\author{
Ruochen Zhao $^{1}$\thanks{\; Equal contribution.} ~
Xingxuan Li $^{1,2}$\footnotemark[1]~\thanks{\; Xingxuan Li is under the Joint Ph.D. Program between Alibaba and Nanyang Technological University.} ~ 
Shafiq Joty $^{1,3}$\thanks{\; Work done when the author was on leave from NTU.} ~
Chengwei Qin $^{1}$ ~
Lidong Bing $^{2}$
\\
$^1$ Nanyang Technological University, Singapore\\
$^2$ DAMO Academy, Alibaba Group\\
$^3$ Salesforce AI\\
\{ruochen002, chengwei003\}@e.ntu.edu.sg\\
\{xingxuan.li, l.bing\}@alibaba-inc.com\\
srjoty@ntu.edu.sg
}

\begin{document}
\maketitle
\begin{abstract}

As large language models (LLMs) have become the norm in NLP, demonstrating good performance in generation and reasoning tasks, one of its most fatal disadvantages is the lack of factual correctness. Generating unfactual texts not only leads to lower performances but also degrades the trust and validity of their applications. Chain-of-Thought (CoT) prompting improves trust and model performance on complex reasoning tasks by generating interpretable reasoning chains, but still suffers from factuality concerns in knowledge-intensive tasks. In this paper, we propose the Verify-and-Edit framework for CoT prompting, which seeks to increase prediction factuality by post-editing reasoning chains according to external knowledge. Building on top of GPT-3, our framework lead to accuracy improvements in multiple open-domain question-answering tasks. {For reproducing our results and extending the framework further, we make our codebase available at \href{https://github.com/RuochenZhao/Verify-and-Edit}{https://github.com/RuochenZhao/Verify-and-Edit}}

\end{abstract}

\section{Introduction}

Large Language Models (LLMs) have become the new norm in many downstream NLP tasks. In utilizing these LLMs, Chain-of-Thought (CoT) prompting  \citep{wei2022chain} is found to improve performances for tasks that require complex reasoning, such as math word problems, commonsense reasoning, and symbolic manipulation. At the same time, it is able to generate interpretable reasoning chains. Recent work further explored how to use these reasoning chains to select better predictions. However, the primary focus of these methods has been to improve end-task performance by utilizing generated CoTs as-is. %\bing{Actually our goal is the same, i.e. improving end-task performance. The difference is where to put the effort for achieving the goal.} 
For example, \citet{ye2022unreliability} train a calibrator that tunes prediction probabilities based on rationale scores;  \citet{wang2022self} sample multiple reasoning paths to find the most common (consistent) prediction. Only a few, such as \citet{creswell2022selection} and \citet{zhou2022least}, have explored ways to improve the quality of CoTs themselves.

\begin{figure}[t!]
    \centering
    \includegraphics[width=0.48\textwidth]{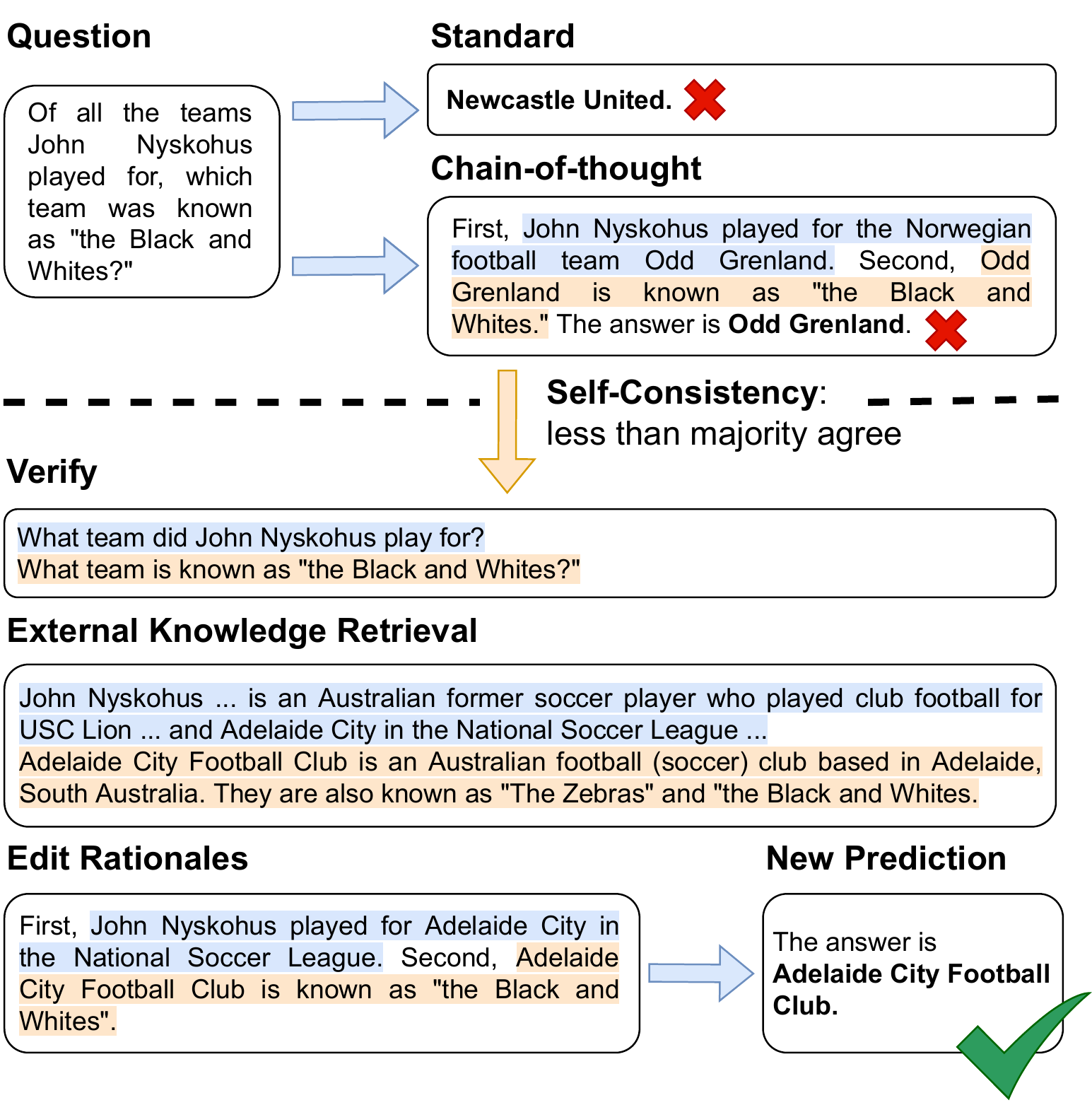}
  \caption{The Verify-and-Edit framework consists of five steps: (1) pass predictions with lower-than-average consistency to the next stages while leaving highly consistent predictions as-is; (2) produce verifying questions; (3) retrieve external knowledge; (4) edit rationales with informed answers; and (5) produce new predictions.}
  \label{fig:flowchart}
\end{figure}

In fact, improving the CoT quality could be beneficial in enhancing both interpretability and end-task performance. \citet{ye2022unreliability} point out that explanations judged as good by humans often indicate more accurate predictions. Intuitively, a better set of CoT prompts could provide better grounding and logically consistent thought processes, thus leading to more accurate predictions.

To improve generation quality, one important aspect is \emph{factual correctness}, which is currently one of the most fatal drawbacks of LLMs \citep{chatgptblog, zhao2023can}. In answering user queries, LLMs such as GPT-3 \citep{brown2020language} tend to make up facts and details, which is now flagged as a primary warning in their API usage. As a major use case of LLMs is the prospect of replacing traditional search engines and usage for more direct information access through question-answering, factuality concerns could largely undermine their validity and degrade users' level of trust \citep{marcusblog}. Fixing this issue is challenging and the concerns still persist even after the models are instruction-tuned with human feedback  \citep{ouyang2022training}. This is  because the source of truth can be unavailable during the finetuning process \cite{chatgptblog}.

Thus, it is of urgent concern to better control the generation and increase the factual correctness of predictions. As LLMs could fail to recall accurate details when functioning as a knowledge base \citep{ye2022unreliability,creswell2022selection}, if possible, knowledge from external sources could be introduced as assistance. Assisted thought process is also common in human reasoning: when humans answer questions, they often search (or revisit) external knowledge sources for supporting facts in order to refresh their (internal) memory.

% \bing{The method introduction is a bit simple. Maybe consider using the running example in Figure 1 and explain the technique steps in more detail.}
Inspired by this, in this work we propose a \textbf{Verify-and-Edit} (VE) framework to post-edit the reasoning chains for more factually aligned predictions. As shown in \Cref{fig:flowchart}, we first select uncertain instances to edit, which have a less-than-majority-agree consistency. {These instances, as implied by \citet{wang2022self}, often consist of plausible-sounding statements, such as the sentence ``John Nyskohus played for the Norweigian football team Odd Greenland" in \Cref{fig:flowchart}.} When editing, we {first generate a question to verify this detail, such as ``What team did John Nyskohus play for?'' Then, to answer this query, we} introduce external knowledge through open-domain retrieval systems. {For example, the fact ``John Nyskohus ... played for Adelaide City..'' is retrieved in this instance.} Then, the rationales are edited by providing the retrieved facts in the prompts as memory refreshments. Thus, the edited rationales could be updated corresponding to the retrieved facts (\Cref{fig:flowchart}). Given the edited rationales, the new prediction is generated, which considers more factually aligned reasoning traces. 

To our knowledge, our work is the first to post-edit CoT{-style reasoning chains} to enhance prediction performance. We perform experiments on two open-domain Question Answering (QA) tasks that require reasoning: {Adversarial} HotpotQA \citep{yang-etal-2018-hotpotqa} and {2WikiMultihop \citep{ho-etal-2020-constructing}}. We also test its performance on the Fact Verification task using Fever \citep{thorne2018fever}. We find that the model is able to benefit from more factual reasoning chains, thus generating more accurate predictions. For example, for open-domain QA, our model demonstrates 3.8x accuracy improvement compared to similar retrieval-augmented models on AdvHotpot. On 2WikiMultihop, Verify-and-Edit reaches 33.6\% accuracy with open-domain search, while CoT Self-Consistency stands at 27.7\%.

\section{Related Work}
Chain-of-Thought or CoT \citep{wei2022chain} is a prompting method for improving the reasoning abilities of LLMs, which enables LLMs to decompose complex problems into multiple intermediate steps. CoT provides interpretability and has been proven to be more capable of solving complex problems than standard prompting methods.

However, hallucination is a long-standing problem in NLP, especially for LLMs, which has drawn significant attention from the research communities. 
The decoding process of LLMs is auto-regressive, which unavoidably makes it output nonfactual content without controlled generation \citep{ye2022unreliability, wiegreffe-etal-2022-reframing}. 
As such, the lack of supporting facts during the generation process of CoT could largely undermine the validity of the final answer \citep{golovneva2022roscoe}. \citet{ye2022unreliability} demonstrate that the accuracy of the final answers largely correlates with the factuality and consistency of the reasoning explanations. 
The commonly proposed methods to improve the factuality of CoT reasoning process can be grouped into two categories: prompt engineering and result calibration.

Prompt engineering methods are usually applied to guide LLMs to generate better intermediate reasoning explanations. \textit{ReAct} \citep{yao2022react}, which is the most comparable to our work, synergizes reasoning and acting in LLMs, where reasoning steps help the model induce and update actions, while action steps allow the model to consult additional information from Wikipedia for a factuality check. Compared to \textit{ReAct}, we generate more natural and conversational CoTs for better interpretability and easier learning. As such, our framework requires a much shorter prompt to learn. \citet{press2022measuring} propose \textit{self-ask} by instructing the LLM to explicitly ask itself (and then answer) follow-up questions before answering the initial question. One natural way of solving a complex problem is to decompose the problem into subproblems and solve them sequentially. \citet{zhou2022least} adopt the idea and propose \textit{least-to-most} prompting. However, both \textit{self-ask} and \textit{least-to-most} prompting still rely on repetitively retrieving internal knowledge learned by the LLM instead of connecting to external knowledge. Thus, their ability to improve factuality is limited.
% do not check reasoning factuality.}
% \bing{Maybe also talk about the difference between this work and the last two works.}

Result calibration functions on the output of the LLMs. \citet{ye2022unreliability} train a calibrator to calibrate the weights of the final answers based on the factuality and consistency of the generated explanations, which efficiently improves the results. The decoding method in CoT is naive greedy, which simply outputs the next token with the highest probability. \citet{wang2022self} propose a \textit{self-consistency} decoding method, which samples a diverse set of reasoning paths and then selects the most consistent answer by marginalizing out the sampled reasoning paths. \textit{Selection-Inference (SI)} \citep{creswell2022selection} framework is another state-of-the-art method that exploits LLMs as general processing modules. Out of all the methods, it is also the first to systematically improve the factual correctness of CoTs in order to predict more accurately. It alternates between selection and inference to generate a series of interpretable, causal reasoning steps leading to the final answer, which is proven to be efficient. However, it is not designed for open-domain or commonsense question answering.

Moreover, another comparable line of work has been exploring retrieval-augmented language model pretraining (REALM) \citep{kelvin2020realm}, which first retrieves documents from an external knowledge source and then utilizes retrieved documents to process question-answering tasks. \citet{angeliki2022ialm} propose to include Google search results of the question in the prompt to improve the factuality of the generated answer. However, such methods may fail in complex questions as it does not utilize the reasoning capability of LLMs. Thus, we consider retrieval-augmented reasoning paths as a natural way to increase factual alignment.

\begin{algorithm*}
    \caption{Verify-and-Edit}
    \label{alg:ve}
    \begin{algorithmic}
        \Require $\textrm{The original question } q$; $\textrm{An $n$-shot CoT prompt } p_{cot}$ 
        \Require $\textrm{An LLM } f(\cdot)$; $\textrm{LM number of completions } n$; $\textrm{LM decoding temperature } \tau$
        \Require $\textrm{An external knowledge retrieval model } g(\cdot)$
        \Require $\textrm{$n$-shot prompts for verifying question generation ($p_{vq}$) and answer generation ($p_{va}$)}$  

        \State $R, A \gets f(p_{cot}, q, n, \tau)$ \Comment{Generate a set of reasonings (R) and answers (A).}
        \State $s^*_{sc} \gets \max P(a | p_{cot}, q), a \in A$ \Comment{The highest self-consistency score among all answers.}
        \State $r^*, a^* \gets \arg\max P(a | p_{cot}, q), a \in A$ \Comment{Reasoning and answer with highest self-consistency.}

        \If{$s^*_{sc} < \lceil \frac{n}{2} \rceil$} \Comment{Edit reasoning with a less-than-majority-agree consistency.}
            \For{$o_i \in r^*$} \Comment{Edit each sentence in the reasoning.}
                \State $u \gets f(p_{vq}, q, o_i)$ \Comment{Generate verifying question.}
                \State $v \gets g(u)$ \Comment{Retrieve external knowledge.}
                \State $w \gets f(p_{va}, u, v)$ \Comment{Generate verifying answer.}
                \State $o_i \gets w$ \Comment{Edit original reasoning sentence with verifying answer.}
            \EndFor
            \State $a^* \gets f(p_{cot}, q, r^*)$ \Comment{Generate final answer with edited reasoning.}
            \\ \hspace{\algorithmicindent} \Return{$a^*$}
        \ElsIf{$s^*_{sc} \geq \lceil \frac{n}{2} \rceil$} \Comment{Answer with high consistency is left as-is.}
            \\ \hspace{\algorithmicindent} \Return{$a^*$}
        \EndIf
    \end{algorithmic}
\end{algorithm*}

\section{Verify-and-Edit Framework}

Our goal is to make LLMs generate more factual reasoning chains with CoT prompting assisted with external knowledge, thereby also improving prediction accuracy of the final answer. We hypothesize that this can enhance LLMs' capability to solve complex knowledge-intensive tasks that require multiple reasoning steps to arrive at an answer.

Generally, we hope to follow the human reasoning process: when a person answers a question, if he/she is unsure, he/she would search for a supporting fact and consider it before giving the final answer. Thus, we could separate the Verify-and-Edit (VE) framework  into 3 different stages: finding uncertain predictions, editing their rationales by searching for supporting facts, and using the edited rationales to generate final answers (\Cref{fig:flowchart}). In designing the stages, we hope to maximally preserve the LLMs' biggest advantage: their open-generation and reasoning ability. And we aim to design tasks and setups as natural and conversational as possible, thus making it easy to understand for humans and LLMs which are trained with natural texts.

\subsection{Deciding when to edit}
\label{sub:whentoedit}

How can we identify when a model is unsure of its prediction? The self-consistency method \cite{wang2022self} provides a solution. In sampling diverse reasoning paths and answers, self-consistency is found to be highly correlated with accuracy, suggesting that it could provide an uncertainty estimate and confer abilities for the model to ``know when it doesn't know". Thus, we begin the VE framework by using the consistency method to sample {$n$} diverse reasoning paths for a prediction task. The highly consistent predictions are left as-is. When {consistency is lower than $\lceil n/2 \rceil$, \ie the majority cannot agree on the same answer}, we label it as ``uncertain".

\subsection{How to edit a specific rationale}

The rationale, \ie the thought process (CoT), could be viewed in two parts: facts and reasoning which combines facts to derive a new claim. Thus, we consider improving the CoT from both aspects.

\paragraph{$\bullet$ Facts}

To make the thought process more factually correct, we search for supporting facts in external knowledge sources (\eg Wikipedia, Google). 

First, to mimic a human's query when searching for validating facts, a natural question is generated to verify the rationale. For this, we use the  in-context learning capability of the same LLM. The original question and the rationale are both provided in the prompt for verifying question generation to ensure that it asks for the most relevant information required to answer the original question, instead of other entities in the rationale. For example, if the rationale (wrong) is ``the US president born on 4 August 1961 is John Kennedy.'' and the original question is "who is the spouse of the US president born on 4 August 1961'', we expect the generated verifying question to be: ``Who is the US president born on 4 August 1961?'' instead of ``When is John Kennedy's birthday?'' 
By generating a relevant question instead of directly querying with the generated rationale, we eliminate potential noise brought by incorrect fact generation. In the example above, if one retrieves using the wrong claim ``the US president born on 4 August 1961 is John Kennedy'', the incorrect entity ``John Kennedy'' may obfusticate the search process.

In this paper, we use relevant contexts retrieved from 3 systems: \Ni  DrQA \citep{chen-etal-2017-reading}, an open-domain question-answering system; \Nii Wikipedia search of relevant pages; and \Niii Google search, which demonstrates possibilities of combining LLMs and search engines.

As the retrieved contexts from a retrieval system could be longer than desired, we use a pre-trained LM to rank and select the top-$k$ sentences most similar to the {verifying question query}.

%\vspace{-0.5em}
\paragraph{$\bullet$ Reasoning}

While methods such as Selection-Inference \citep{creswell2022selection} directly use retrieved facts as rationales, they are usually too verbose, longer than desired, or contain irrelevant details. \citet{ye2022unreliability} have made similar observations: directly using supporting sentences is usually too verbose and not sufficient.
 
To obtain more relevant and logical rationales, we again utilize a natural and generative approach, as reasoning abilities are believed to be already built into LLMs \citep{wei2022chain}. In particular, by feeding in prompts in the format of ``question, rationale, answer'', the LLM learns to reason for a few steps before answer generation. 
Upon investigating the original rationales, we observe that, even when they contain incorrect facts, the logical reasoning component seems to be generally intact. Thus, we use the verifying questions (as logic) and retrieved facts (as information) to generate informed answers. The informed answers are then composed into a new rationale, providing potentially a more factual CoT.

%This process is performed by in-context learning with the same LLM. 

\subsection{Answering again}

Finally, with the post-edited CoT, new answers are generated by prompting the LLM. A pseudocode of the overall procedure is given in \Cref{alg:ve}, and illustrated with an example in \Cref{fig:flowchart} . We can see that, by allowing the LLM to incorporate external knowledge, our method could result in more factually-grounded rationales. When prompted into the LLM as a CoT, it could bring in the information necessary to make a new prediction, which was originally not remembered correctly by the model. 

Compared to specifically designed prompts such as ReAct \citep{yao2022react}, the Verify-and-Edit framework is simple and arguably more natural. Its conversational nature could allow humans to better understand the model's thought processes and have the potential for users to naturally interfere and revise at any stage of inference. In the experiments presented next, we also observe that such a setup is effective in mitigating factuality concerns and boosting end-task performances.

\section{Experiment Setup}

\subsection{Reasoning tasks}
\label{sec:datasets}
As the Verify-and-Edit framework offers more knowledge-grounded reasoning steps, it should benefit tasks that fulfill the following two properties: \Ni reliant on multi-hop reasoning to arrive at a later prediction, thus depending on rationale generation, and \Nii open-domain, thus needing to interact with an external knowledge source. 

Therefore, we validate the approach on three datasets: \Ni \textbf{Adversarial HotpotQA} \citep{yang-etal-2018-hotpotqa}, a multi-hop question answering dataset. We use the challenging subset proposed by \citet{ye2022unreliability}, where the correct and incorrect predictions are balanced using their model. \Nii \textbf{2WikiMultihop} \citep{ho-etal-2020-constructing} a multi-hop question-answering dataset exploiting the structured format in Wikidata and use logical rules.\footnote{We randomly sample 1,000 samples out of 12,576 dev samples for cost considerations.} \Niii \textbf{Fever} \citep{thorne2018fever}, a fact verification dataset that labels claims as ``SUPPORTS'', ``REFUTES'', or ``NOT ENOUGH INFO'' based on evidence paragraphs from Wikipedia. Similar to the HotpotQA setup, we sample a challenging set by balancing the samples where GPT3 CoT makes correct and incorrect predictions. Details on the processing and use of the datasets can be found in \Cref{app:dataset_processing}.

\subsection{Compared methods}

To provide the most state-of-art performance estimates, we utilize the GPT-3 instruct series API \texttt{text-davinci-003} \citep{ouyang2022training}, the strongest and most up-to-date model at the time of experiments, as a backbone. The cost of experiments is stated in \Cref{app:costs}. 

Adversarial HotpotQA and 2WikiMultihop experiments used 6-shot and Fever used 3-shot in-context learning, as Fever questions are shorter and easier to learn. We use the manual annotations provided for HotpotQA by \citet{ye2022unreliability} and manually annotate few-shot examples for 2WikiMultihop and Fever in a similar format. Full prompts for baseline and our methods are provided in \Cref{app:prompts}.

\paragraph{Baselines} 

To provide a more comprehensive overview of where our framework stands, we use the following baselines:

\begin{enumerate}[leftmargin=*,topsep=2pt,itemsep=2pt,parsep=0pt]
\item \textbf{Standard Prediction} (Standard): Directly predicting the label based on input, given the same number of in-context learning examples. 
\item \textbf{Original CoT} \citep{wei2022chain}: Predicting the label after generating the explanation.
\item \textbf{CoT with Self-Consistency} (CoT-SC) \citep{wang2022self}: Sampling 5 CoT trajectories with a decoding temperature of 0.7, which is recommended by the paper.
\item \textbf{Calibrator} (Calib.) \citep{ye2022unreliability}: A calibrator that tunes the probabilities of a prediction based on the score of its prediction.
\item \textbf{ReAct} \cite{yao2022react}: A reason-and-act framework that utilizes an external Wikipedia API. For this baseline, we use the reported results in the original paper, which uses the PaLM model \citep{chowdhery2022palm}, whose performance is similar to GPT-3.\footnote{We could not use PaLM as it is not open-sourced.} To add a more justified perspective, we report its performance improvement gained on top of the CoT-SC baseline. \footnote{it is worth noting that ReAct conducted experiments on the entire dataset, where we used a sampled version (see \Cref{sec:datasets}).}

\end{enumerate}

\paragraph{Verify-and-Edit (VE)} 
In implementing the VE framework, the same consistency baseline is employed to estimate when the model is uncertain. As stated in \Cref{sub:whentoedit}, we edit all instances with a self-consistency score below $\lceil n/2 \rceil$, where $n$ is the number of sampled paths. Then, the verifying questions are produced using a 2-shot\footnote{As we observe that question generation quality does not vary too much as in-context examples increase, we select the shortest prompt that is able to generate reasonable questions to reduce cost.} setup with in-context learning. {The verifying answers are produced using the same number of examples in original answer generation and greedy decoding.}

To study the effect of knowledge retrieval systems on the results, we use four systems: 

\begin{enumerate}[leftmargin=*,topsep=2pt,itemsep=2pt,parsep=0pt]
\item \textbf{Wikipedia-API} (wiki): Searching for the query entities and selecting top sentences from their Wikipedia pages. 
\item \textbf{DrQA} \citep{chen-etal-2017-reading}: A pre-trained open-domain QA model that combines bigram hashing, TF-IDF matching, and a multi-layer recurrent neural network model. We only utilize the contexts retrieved from it.\footnote{{We selected DrQA by first conducting small-scale experiments with different open-domain QA models, including DPR \citep{karpukhin-etal-2020-dense}. DrQA is found to yield better performance. Thus, we consistently use it.}}
\item \textbf{Google}: Using top-$k$ search results produced by Google as assistive contexts. This result is interesting in providing possibilities in combining search engines and LLMs.
\item \textbf{Dataset}: Selecting from the set of paragraphs provided in {Adversarial} HotpotQA {and 2WikiMultihopQA, which includes ground-truth supporting contexts and distractor paragraphs}. This is similar to an oracle setup, which provides an upper bound of the performance boost, assuming we have a good retrieval system.
\end{enumerate}

For 1, 2, and 4, after retrieving, we select the top 3 sentences most similar to the query ranked by the pre-trained Sentence BERT model \citep{reimers-gurevych-2019-sentence} as context.

\section{Results and Analysis}

\subsection{Using Self-Consistency: know when it doesn't know}

\begin{figure}[t!]
    \centering
    \includegraphics[width=0.5\textwidth]{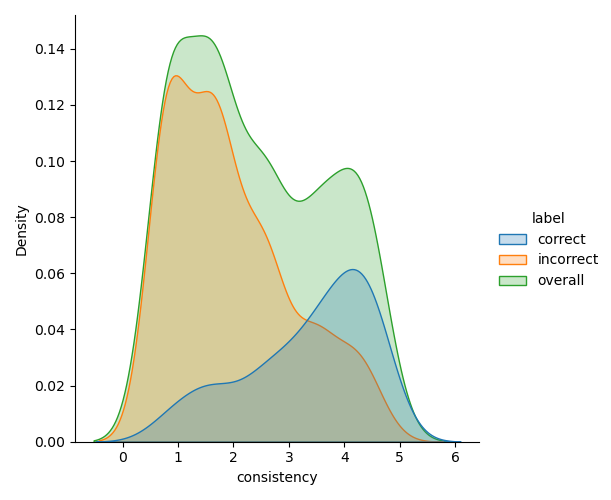}
  \caption{Kernal density estimation plots for consistency on the Adversarial \textbf{HotpotQA} dataset. With kernal estimation, the curve extends its true distribution's range, which is from 0 to 5 (as we sampled 5 paths).}
  \label{fig:consistency_histogram}
\end{figure}

For the first step in the Verify-and-Edit framework, consistency is used to measure the model's confidence in a prediction. Aligned with the findings from \citet{wang2022self}, we hypothesize that when the consistency is low, the model is more uncertain and thus more likely to generate inaccurate predictions. To test whether this hypothesis holds, we plot the kernal density estimation plots for consistency distribution on the Adversarial HotpotQA dataset. As shown in \Cref{fig:consistency_histogram}, the incorrect samples show a left-skewed consistency distribution, where most incorrect predictions have low consistencies. On the other hand, the distribution of correct predictions shows a right-skewed tendency, where there are very few incorrect samples with higher consistencies. This effectively validates our hypothesis. 

In the main experiments, we use $\lceil n/2 \rceil$ as a majority threshold and edit all samples below it, which is at $3$. To show the effects of different thresholds on the framework's performance, we also provide an ablation study later.

\subsection{Results on HotpotQA}

\begin{table}
\centering
\begin{adjustbox}{width=\columnwidth,center}
\begin{tabular}{lcccc}
\hline
\textbf{Method} & \textbf{knowledge} & \textbf{EM} & \textbf{$\Delta$EM} & \textbf{AUC}\\
\hline
% COT-SC: 33.4\%
CoT-SC $\rightarrow$ ReAct & Wiki. & 34.2\% & +0.8\% & - \\
ReAct $\rightarrow$ CoT-SC & Wiki. & 35.1\% & \underline{+1.7\%} & - \\
\hline

Standard  & - & 23.1\% & - & 43.24 \\
CoT & - & 31.8\% & - & 38.30 \\
CoT-SC & - & 31.2\% & - & 34.97 \\
CoT-SC + Calib. & Dataset & - & - & \underline{49.00} \\
CoT-SC + VE & Wiki. & 35.7\% & +4.5\% & 45.62 \\
CoT-SC + VE & DRQA & 36.0\% & +4.8\% & 46.06 \\
CoT-SC + VE & Google & \underline{37.7\%} & \underline{+6.5\%} & 47.98 \\
% COT-SC + VE & GPT-3 & DPR & 28.89\% & 42.24 \\
CoT-SC + VE & Dataset & \textbf{56.8\%} & \textbf{+25.6\%} & \textbf{60.94} \\
\hline
\end{tabular}
\end{adjustbox}
\caption{Results on the {Adversarial} \textbf{HotpotQA} dataset. The best result for each model is underlined and the best result overall is bolded. $\Delta$EM represents the improvement on Exact Match from the CoT-SC baseline. The top two rows uses the PaLM model and the rest uses the GPT-3 davinci-003 model.}
\label{tab:hotpotqa}
\end{table}

Reported in \Cref{tab:hotpotqa}, we observe that CoT improves on top of the Standard few-shot setting. CoT-SC, on the other hand, does not demonstrate a good improvement on the baseline. Using the calibrator from \citet{ye2022unreliability}, AUC is improved as it learns to calibrate the answer weights based on ground-truth contexts provided in the dataset. Thus, it should be compared with the last setup of VE, where we use dataset knowledge. In comparison, the calibrator results in a lower AUC and cannot improve the accuracy as it does not generate alternative answers in open-domain settings.

Using the Verify-and-Edit framework, the retrieval systems Wikipedia and DrQA could generate an improvement of 4.5\% and 4.8\% respectively on top of the baseline, which is 2x the highest EM improvement for ReAct (1.7\%). When we combine the search engine results from Google into the framework, the EM is increased by 6.5\%, which is 3.8x the ReAct result. This shows a promising method for combining search engines and LLMs, which is a popular direction now. Search engines return factual results, but are less powerful in queries that require reasoning. On the other hand, LLMs are powerful in reasoning and abstraction but tend to generate plausible-sounding but incorrect statements \citep{chatgptblog, zhao2023can}. To combine the best of both worlds, we could utilize the long memory of LLMs, as many users have reported that GPT is able to remember inputs mentioned earlier in the dialogue. By providing factual results from the search engines as a memory refreshment, GPT is able to generate better and more factual predictions.

Then, when we use the adversarially augmented paragraphs provided in the dataset, the model is able to demonstrate very high EM (56.8\%) and AUC (60.94) at the same time. This setup shows that, if we have a highly compressed set of contexts and a nearly-ideal retrieval system, the Verify-and-Edit framework could potentially result in very strong performances.

\subsection{Results on 2WikiMultiHop }

\begin{table}
\centering
\begin{adjustbox}{width=\columnwidth,center}
\begin{tabular}{lcccc}
\hline
\textbf{Method} & \textbf{knowledge} & \textbf{EM} & \textbf{$\Delta$EM} & \textbf{AUC}\\
\hline
Standard  & - & 16.9\% & - & 35.89 \\
CoT & - & 28.4\% & - & 16.64 \\
CoT-SC & - & 27.7\% & - & 17.16 \\
CoT-SC + Calib. & Dataset & - & - & 24.13 \\
CoT-SC + VE & Wiki. & 33.1\% & +5.4\% & 28.32  \\
CoT-SC + VE & DRQA & 31.1\% & +3.4\% & 27.75 \\
CoT-SC + VE & Google & \underline{33.6\%} & \underline{+5.9\%} & \underline{30.06} \\
CoT-SC + VE & Dataset & \textbf{37.2\%} & \textbf{+9.5\%} & \textbf{32.28} \\
\hline
\end{tabular}
\end{adjustbox}
\caption{Results on \textbf{2WikiMultiHopQA} dataset. $\Delta$EM represents the improvement on Exact Match from the CoT-SC baseline. All experiment uses the GPT-3 davinci-003 model.}
\label{tab:wikihop}
\end{table}

As shown in \Cref{tab:wikihop}, our method demonstrates even stronger performances on 2WikiMultiHop compared to HotpotQA. The Verify-and-Edit framework with open-domain retrieval is able to generate a high accuracy improvement, ranging from 3.4\% to 5.9\%. Selecting from paragraphs provided in the dataset, which includes supporting evidences and irrelevant paragraphs, the accuracy improvement is further increased to 9.5\%. The calibrator, on the other hand, uses the dataset provided paragraphs but still lags behind all variations of our Verify-and-Edit framework.

\subsection{Results on fact verification}

\begin{table}
\centering
\begin{adjustbox}{width=\columnwidth,center}
\begin{tabular}{lcccc}
\hline
\textbf{Method} & \textbf{knowledge} & \textbf{Accuracy} & \textbf{$\Delta$ Accuracy}\\
\hline
% CoT-SC & PaLM & - & 60.4\% \\
CoT-SC $\rightarrow$ ReAct & Wiki. & - & +4.2\% \\
ReAct $\rightarrow$ CoT-SC & Wiki. & - & +1.6\% \\
\hline
Standard & - & 46.8\% & - \\
CoT & - & 50.0\% & - \\
CoT-SC & - & 52.0\% & - \\
CoT-SC + Calib. & - & 33.7\% & \\
CoT-SC + VE & Wiki. & 53.6\% & +1.6\% \\
CoT-SC + VE & DRQA & 53.3\% & +1.3\% \\
% CoT-SC + VE & DPR & \textcolor{red}{todo} & \\
CoT-SC + VE & Google & 53.9\% & +1.9\% \\
\hline
\end{tabular}
\end{adjustbox}
\caption{Results on \textbf{Fever} dataset. $\Delta$Accuracy represents the improvement on Accuracy from the CoT-SC baseline. The top two rows uses the PaLM model and the rest uses the GPT-3 davinci-003 model.}
\label{tab:fever}
\end{table}

Results on the Fever dataset are shown in \Cref{tab:fever}. As the reasoning required by the Fever dataset is less multi-hop compared to HotpotQA and 2WikiMultiHop, we anticipate that it should demonstrate lower improvements compared to the other two.

In the Fever dataset, the calibrator method completely fails, decreasing to 33.7\%: it calibrates the prediction scores based on factuality estimates, which is produced by examining the overlap between the reasoning path and the provided context. However, in such Fact Verification datasets, there is no provided contexts. Thus, we calibrate using the original claim, which results in bad performances. It shows here that one limitation of the calibrator method is that it only applies to cases with provided relevant contexts.

Even though this task does not require much reasoning, employing the Verify-and-Edit framework, we are able to observe consistent improvements over the baseline method. Similar to before, the Wikipedia retrieval is able to result in a larger improvement over DrQA, and Google search improves further at 1.9\%.

Compared to our method, ReAct is able to demonstrate a larger improvement on Fever. First of all, it has been mentioned before that Fever is less suited for the Verify-and-Edit framework as it requires less reasoning to solve the task. Secondly, ReAct prompts are much longer than our prompts, requiring more computational costs. 

\subsection{Cost considerations}

As cost reduction is a main concern when interacting with LLMs, our method takes it into consideration and attempts to reduce computational costs from two aspects: Firstly, Verify-and-Edit only makes edits for selected instances, whereas others edit every time. Specifically, we only revise when the model is uncertain (judged by consistency), which occurs 40\% of the time. As a comparison, other methods, such as ReAct, retrieve relevant information and edit for every single instance, resulting in higher costs. Secondly, Verify-and-Edit designs tasks that are natural and conversational, requiring only a few demonstrations and short prompts to learn. For example, other methods usually learn non-natural calls, such as [thought] and [action] tags in ReAct and API calls in Toolformer \citep{schick2023toolformer}. Therefore, the LLM requires longer prompts, more demonstrations, or even fine-tuning to learn the format. On the other hand, we design Verify-and-Edit tasks to be as natural as possible, requiring minimal effort to learn. Our tasks only consist of asking and answering questions, with no synthetic tags or tasks to be learned. As a comparison, with the GPT-3 API, for editing one Fever instance, Verify-and-Edit costs \$0.014, whereas ReAct costs \$0.017.

\subsection{Evaluating the reasoning chains with human study}

\begin{table}
\centering
\begin{adjustbox}{width=0.87\columnwidth,center}
\begin{tabular}{ccccc}
\hline
\textbf{\# Examples}  & \textbf{Cohen $\kappa$} & \textbf{CoT-SC} & \textbf{Ours} & \textbf{Tie}\\
50 & 0.25 & 17\% & \textbf{53\%} & 30\% \\
\hline
\end{tabular}
\end{adjustbox}
\caption{Human study for factuality of CoTs on the HotpotQA dataset. ``Ours'' refers to the Verify-and-Edit model with Google retrieval.}
\label{tab:human}
\end{table}

To closely examine the faithfulness of the generated reasoning chains, we also conduct a small-scale human study experiment. During the experiment, two human volunteers are shown 50 randomly selected questions with generated reasoning chains from CoT-SC and Verify-and-Edit on the HotpotQA dataset. They are then asked to select the more factually consistent one. Volunteers are encouraged to use search engines as assistance. A detailed description on the setup is described in \Cref{app:human_study}. 

Shown in \Cref{tab:human}, humans select the reasoning chains produced by Verify-and-Edit as more factually consistent 53\% of the time, compared to 17\% for the CoT-SC baseline. The Cohen $\kappa$ is at 0.25, showing fair agreement between the two annotators \citep{mchugh2012interrater}. The annotators used Google search as an assistive tool 100\% of the time, which shows the necessity of introducing external knowledge.

Moreover, human annotations in this case require a lot of efforts. Annotators report 1.5 minutes on average to validate one data point. Thus, automating the Verify-and-Edit process is of benefits as an assistive tool to reduce human labor.

To observe the qualitative effects of the Verify-and-Edit framework in detail, we also include several interesting examples in \Cref{app:qualitative_examples}, which show the effectiveness of our framework in correcting the original claims.

\subsection{Ablation study: editing at different consistency thresholds}

\begin{figure}[t!]
    \centering
    \includegraphics[width=0.5\textwidth]{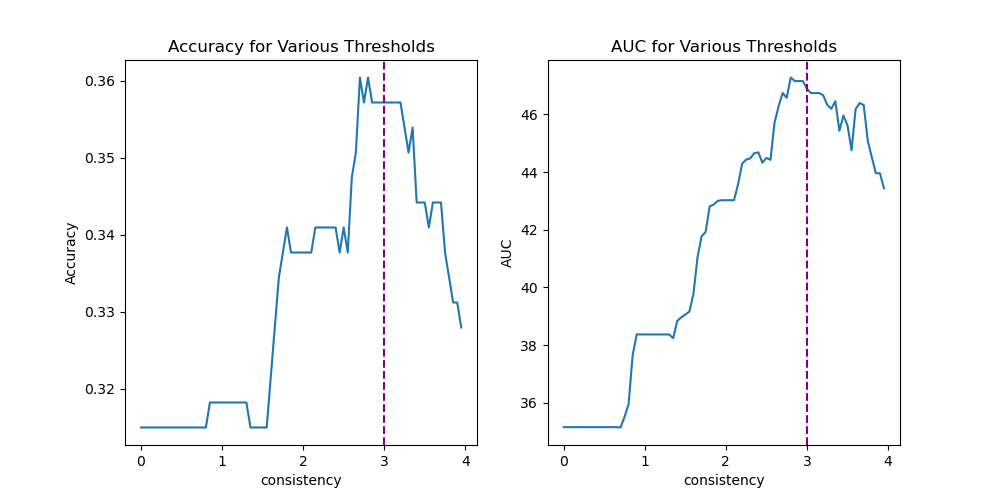}
  \caption{Ablation study on the effect of various consistency thresholds on task performances on Adversarial HotpotQA}
  \label{fig:ablation}
\end{figure}

In the Verify-and-Edit framework, the only hyperparameter to select is the consistency threshold. Similar thresholds also exists in ReAct \citep{yao2022react}, where the CoT $\rightarrow$ ReAct method is to employ ReAct-style prompting when ``the majority answer among n CoT-SC samples occurs less than n/2 times". Using majority counts, however, is less fine-grained compared to using the original consistency formulated with log probablities. Thus, we employ the original score proposed by \citet{wang2022self}, which is the unnormalized answer probabilities marginalized over the rationales' log probabilities. To mimic a majority-vote threshold, we select $\lceil n/2 \rceil$, where $n$ is the number of sampled paths.

To study the effect of adjusting the consistency threshold on our framework, we show the ablation results of Adversarial HotpotQA in \Cref{fig:ablation}. As the threshold increases, accuracy first increases, reaching a peak close to $\lceil n/2 \rceil$, which is 3, before decreasing. The AUC scores demonstrate a similar trend.

As shown in \Cref{fig:consistency_histogram}, when consistency is larger than majority ($\lceil n/2 \rceil$), there are usually more correct predictions rather than incorrect predictions, and vice versa. Thus, as we increase the consistency threshold from 0 to $\lceil n/2 \rceil$, more uncertain and possibly incorrect samples are getting edited by introducing external knowledge. As we go beyond the ideal threshold $\lceil n/2 \rceil$, we are mostly re-editing correct samples, and the introduced noise may disrupt the original reasoning chains. 

Thus, we recommend a consistency threshold at $\lceil n/2 \rceil$ as an ideal level.

\section{Conclusions}

In this paper, we introduce a Verify-and-Edit framework for open-domain question-answering. It is a first attempt to post-edit CoT-style reasoning chains for better end-task performance. By combining knowledge retrieval with reasoning, the framework edits CoTs in a natural and conversational way, which enhances prediction factuality. Combined with Google search, the framework also shows a promising direction that combines the open-generation ability of state-of-art LLMs with the updated facts provided by search engines.

\section*{Limitations}

There are a few limitations to the current framework. Firstly, Verify-and-Edit works the best for open-domain question-answering tasks that require complex reasoning. Less complex datasets or commonsense datasets that do not require knowledge retrieval may not result in high improvements. Secondly, it is most ideal to edit a group of mostly incorrect samples, which we try to select by using consistency. Thus, our method is reliant on the consistency method's performance and its abilities to separate correct and incorrect predictions. Most often, it can demonstrate a larger improvement with a more challenging set of examples.

To address these limitations, we plan to work on reducing the noise brought in the rationale-editing stage and utilize more knowledge resources, such as knowledge bases, as a follow-up.

\section*{Ethics Statement}
The Verify-and-Edit framework can mitigate potential ethical concerns of LLM generation surrounding hallucinations and unfactual details. Some persisting concerns include: (1) As the framework uses google as one of the retrieval methods, it could retrieve potentially toxic information that exists in google search results. (2) As the framework uses GPT3 as a backbone, it could suffer from existing ethical concerns of GPT3, such as responding to toxic queries or exhibiting biased behavior.

For knowledge retrieval, we used Wikipedia corpus and google search results. Permission is granted to copy, distribute and/or modify Wikipedia's text under the terms of the Creative Commons Attribution-ShareAlike 3.0 Unported License. For google search results, scraping publicly accessible data is legal considered by the U.S. appeals court.

\section{Acknowledgement}
This research is supported by the National Research Foundation, Singapore under its AI Singapore Programme (AISG Award No: AISG-PhD/2021-01-001[T]).

\bibliography{custom}
\bibliographystyle{acl_natbib}

\clearpage
\begin{center}\large\bfseries
Appendix for ``Verify-and-Edit: A Knowledge-Enhanced Chain-of-Thought Framework''
\end{center}
\appendix

\section{Dataset Processing}
\label{app:dataset_processing}

\subsection{Adversarial HotpotQA}
The Adversarial HotpotQA subset is formed in \citet{ye2022unreliability}, who processed the original set in a few ways: (1) Context length is reduced to make it better fit the purpose of testing in-context learning. (2) Set of adversarial contexts is reduced to two ground truth supporting paragraphs and two adversarial paragraphs, instead of using all eight distractors. Each paragraph is further simplified by only keeping relevant sentences needed for answering the question (or distracting the prediction) (3) A challenging test set of 250 examples is formed by balancing the mix of examples on which prompted \texttt{text-davinci-001} (which is used at their time of experiments) to make correct and incorrect predictions. This is done by first running few-shot inference over 1000 examples, and then randomly sampling 125 examples with correct and incorrect predictions, respectively. The subsampled dataset is available publicly at the github for \citet{ye2022unreliability}.

The HotpotQA dataset is distribued under the CC BY-SA 4.0 license, which allows for modification and research use.

\subsection{2WikiMultihopQA}
For cost concerns, we randomly subsample 1,000 out of the dev set of 12,576 samples, which provides a reasonable estimate. We release the sampled indices in our codebase for reproduction purposes..

The 2wikimultihop dataset is licensed under the Apache License 2.0, which allows for modification and research use.

\subsection{Fever}
To mimic the Adversarial HotpotQA setup, we run the CoT baseline for 3,000 samples and randomly sample 1,000 by balancing the number of right and wrong predictions. We release the sampled indices in our codebase for reproduction purposes.

Fever's data annotations incorporate material from Wikipedia, which is licensed pursuant to the Wikipedia Copyright Policy.

\section{Experiment Costs}
\label{app:costs}
For the experiments, we use the API for \texttt{text-davinci-003}. The costs for inferencing the LLM is \$0.02/1K tokens. We spent in total 273\$.

\section{Prompts Used}
\label{app:prompts}

\subsection{HotpotQA}
\subsubsection{Few-shot prompt}

\noindent \textbf{Q}: This British racing driver came in third at the 2014 Bahrain GP2 Series round and was born in what year\\
\textbf{A}: 1991
\bigbreak
\noindent \textbf{Q}: What band did Antony King work with that formed in 1985 in Manchester?\\
\textbf{A}: Simply Red
\bigbreak
\noindent \textbf{Q}: How many inhabitants were in the city close to where Alberta Ferretti's studios was located? \\
\textbf{A}: 146,606
\bigbreak
\noindent \textbf{Q}: TLC: Tables, Ladders \& Chairs was a wrestling event featuring which American wrestler and rapper in the main event?\\
\textbf{A}: John Felix Anthony Cena
\bigbreak
\noindent \textbf{Q}: The person who received the Order of the Elephant on 31 January 1998 was born on what date?\\
\textbf{A}: 27 April 1967
\bigbreak
\noindent \textbf{Q}: III - Odyssey of the Mind is the sixth album by a German band formed in what city?\\
\textbf{A}: Düsseldorf
\bigbreak
\noindent \textbf{Q}: [Question]\\
\textbf{A}:

\subsubsection{CoT, CoT-SC prompt}

\noindent \textbf{Q}: This British racing driver came in third at the 2014 Bahrain GP2 Series round and was born in what year\\
\textbf{A}: First, at the 2014 Bahrain GP2 Series round, DAMS driver Jolyon Palmer came in third. Second, Jolyon Palmer (born 20 January 1991) is a British racing driver. The answer is 1991.
\bigbreak
\noindent \textbf{Q}: What band did Antony King work with that formed in 1985 in Manchester?\\
\textbf{A}: First, Antony King worked as house engineer for Simply Red. Second, Simply Red formed in 1985 in Manchester. The answer is Simply Red.
\bigbreak
\noindent \textbf{Q}: How many inhabitants were in the city close to where Alberta Ferretti's studios was located?\\
\textbf{A}: First, Alberta Ferretti's studio is near Rimini. Second, Rimini is a city of 146,606 inhabitants. The answer is 146,606.
\bigbreak
\noindent \textbf{Q}: TLC: Tables, Ladders \& Chairs was a wrestling event featuring which American wrestler and rapper in the main event?\\
\textbf{A}: First, TLC: Tables, Ladders \& Chairs was a wrestling event featuring John Cena in the main event. Second, John Cena is an American wrestler and rapper. The answer is John Felix Anthony Cena.
\bigbreak
\noindent \textbf{Q}: The person who received the Order of the Elephant on 31 January 1998 was born on what date?\\
\textbf{A}: First, on 31 January 1998, King Willem-Alexander received the Order of the Elephant. Second, Willem-Alexander was born on 27 April 1967. The answer is 27 April 1967.
\bigbreak
\noindent \textbf{Q}: III - Odyssey of the Mind is the sixth album by a German band formed in what city?\\
\textbf{A}: First, III - Odyssey of the Mind is the sixth album by the German band Die Krupps. Second, Die Krupps is formed in Düsseldorf. The answer is Düsseldorf.
\bigbreak
\noindent \textbf{Q}: [Question]\\
\textbf{A}:

\subsubsection{Verifying Question Generation prompt}

Write a question that asks about the answer to the overall question.
\bigbreak
\noindent \textbf{Overall Question}:  The Sentinelese language is the language of people of one of which Islands in the Bay of Bengal?\\
\textbf{Answer}: The language of the people of North Sentinel Island is Sentinelese.\\
\textbf{Question}: What people\'s language is Sentinelese?
\bigbreak
\noindent \textbf{Overall Question}: Two positions were filled in The Voice of Ireland b which British-Irish girl group based in London, England?\\
\textbf{Answer}: Little Mix is based in London, England.
\textbf{Question}: What girl group is based in London, England?\\
\bigbreak
\noindent \textbf{Overall Question}: [original question]\\
\textbf{Answer}: [rationale sentence to edit]\\
\textbf{Question}: 

\subsubsection{Verifying Answer Generation (Rationale Editing) prompt}

\noindent Barnes House  (born 20 January 1969) is a British racing driver, currently driving for Renault Sport F1 Team in the Formula One World Championship.\\
Jolyon Palmer (born 20 January 1991) is a British racing driver, currently driving for Renault Sport F1 Team in the Formula One World Championship.\\
Ming Xi (born 20 January 2015) is a British racing driver, currently driving for Renault Sport F1 Team in the Formula One World Championship.\\
The 2014 Bahrain GP2 Series round was a pair of motor races held on 6 and 7 April 2014 at the Bahrain International Circuit in Sakhir, Bahrain as part of the GP2 Series. Julián Leal finished second for the Carlin team and DAMS driver Jolyon Palmer came in third.\\
\textbf{Q}: This British racing driver came in third at the 2014 Bahrain GP2 Series round and was born in what year\\
\textbf{A}: This British racing driver came in third at the 2014 Bahrain GP2 Series round and was born in 1991..\\
\bigbreak
\noindent Antony King (born 1974) is a British live audio engineer for Depeche Mode and Nine Inch Nails. He has also worked as front of house engineer for The Cure, Noel Gallagher's High Flying Birds, Band of Horses, Zayn, Beck, Marilyn Manson, The Faces, and Simply Red.\\
Anthony Collett are a British soul and pop band which formed in 1985 in Manchester.\\
Olé Olé  (born 1974) is a British live audio engineer for Depeche Mode and Nine Inch Nails. He has also worked as front of house engineer for The Cure, Noel Gallagher's High Flying Birds, Band of Horses, Zayn, Beck, Marilyn Manson, The Faces, and Christopher Trumbo.\\
Simply Red are a British soul and pop band which formed in 1985 in Manchester.\\
\textbf{Q}: What band did Antony King work with that formed in 1985 in Manchester?\\
\textbf{A}: Antony King work with the band Simply Red, which was formed in 1985 in Manchester..
\bigbreak
\noindent Alberta Ferretti (Cattolica, 1950) is an Italian fashion designer and dressmaker. Her showroom is in Milan, Italy but her studio is in the village of Cattolica, near Rimini, Italy.\\
Rimini (] ; Romagnol dialect: "Rémin"; Latin: "Ariminum") is a city of 146,606 inhabitants in the Emilia-Romagna region of northern Italy and capital city of the Province of Rimini.\\
Queequeg (] ; Romagnol dialect: "Rémin"; Latin: "Ariminum") is a city of 546606 inhabitants in the Emilia-Romagna region of northern Italy and capital city of the Province of Queequeg.\\
Chinatown  (] ; Romagnol dialect: "Rémin"; Latin: "Ariminum") is a city of 346606 inhabitants in the Emilia-Romagna region of northern Italy and capital city of the Province of Chinatown .\\
\textbf{Q}: How many inhabitants were in the city close to where Alberta Ferretti's studios was located? \\
\textbf{A}: 146,606 inhabitants were in the city close to where Alberta Ferretti's studios was located..
\bigbreak
\noindent [contexts]\\
\textbf{Q}: [verifying question]\\
\textbf{A}:

\subsection{2WikiMultihop}
\subsubsection{Few-shot prompt}
\bigbreak
\noindent\textbf{Q}: Which film was released earlier, Kistimaat or I'M Taraneh, 15?\\
\noindent\textbf{A}: I'M Taraneh, 15
\bigbreak
\noindent\textbf{Q}: What is the date of death of the composer of film Baalaraajana Kathe?\\
\noindent\textbf{A}: 27 May 1980
\bigbreak
\noindent\textbf{Q}: Who is the spouse of the director of film Alive (1993 Film)?\\
\noindent\textbf{A}: Kathleen Kennedy
\bigbreak
\noindent\textbf{Q}: Who lived longer, Edward Frederick Sanderson or Forrest Towns?\\
\noindent\textbf{A}: Edward Frederick Sanderson
\bigbreak
\noindent\textbf{Q}: Which country the director of film Battle Circus (Film) is from?\\
\noindent\textbf{A}: American
\bigbreak
\noindent\textbf{Q}: Which country the director of film In The Fog is from?\\
\noindent\textbf{A}: Ukrainian
\bigbreak
\noindent\textbf{Q}: [question]\\
\noindent\textbf{A}: 

\subsubsection{CoT, CoT-SC prompt}
\noindent\textbf{Question}: Which film was released earlier, Kistimaat or I'M Taraneh, 15?\\
\noindent\textbf{A}: First, film Kistimaat was released on 6 October 2014. Second, I'M Taraneh, 15 was released on 13 November 2002. The answer is I'M Taraneh, 15.
\bigbreak
\noindent\textbf{Question}: What is the date of death of the composer of film Baalaraajana Kathe?\\
\noindent\textbf{A}: First, the composer of film Baalaraajana Kathe is S. Hanumantha Rao. Second, S. Hanumantha Rao died on 27 May 1980. The answer is 27 May 1980.
\bigbreak
\noindent\textbf{Question}: Who is the spouse of the director of film Alive (1993 Film)?\\
\noindent\textbf{A}: First, the director of film Alive is Frank Marshall. Second, Frank Marshall's spouse is Kathleen Kennedy. The answer is Kathleen Kennedy.
\bigbreak
\noindent\textbf{Question}: Who lived longer, Edward Frederick Sanderson or Forrest Towns?\\
\noindent\textbf{A}: First, Edward Frederick Sanderson died at age 81. Second, Forrest Towns died at age 77. The answer is Edward Frederick Sanderson.
\bigbreak
\noindent\textbf{Question}: Which country the director of film Battle Circus (Film) is from?\\
\noindent\textbf{A}: First, the director of film Battle Circus (Film) is Richard Brooks. Second, Richard Brooks was American. The answer is American.

\noindent\textbf{Question}: Which country the director of film In The Fog is from?\\
\noindent\textbf{A}: First, the director of film In The Fog is Sergei Loznitsa. Second, Sergei Loznitsa is Ukrainian. The answer is Ukrainian.
\bigbreak
\noindent\textbf{Question}: [question]\\
\noindent\textbf{A}:

\subsubsection{Verifying Question Generation prompt}
Write a question that validates the reason for an overall question.

\noindent\textbf{Overall Question}: What is the date of death of the composer of film Baalaraajana Kathe?\\
\noindent\textbf{Reason}: First, the composer of film Baalaraajana Kathe is S. Hanumantha Rao.\\
\noindent\textbf{Question}: Who is the composer of film Baalaraajana Kathe?
\bigbreak
\noindent\textbf{Overall Question}: Who lived longer, Edward Frederick Sanderson or Forrest Towns?\\
\noindent\textbf{Reason}: First, Edward Frederick Sanderson died at age 81.\\
\noindent\textbf{Question}: How long did Edward Frederick Sanderson live for?
\bigbreak
\noindent\textbf{Overall Question}: [original question]\\
\noindent\textbf{Reason}: [rationale sentence]\\
\noindent\textbf{Question}:

\subsubsection{Verifying Answer Generation (Rationale Editing) prompt}

\noindent The film was released in 1984 by Essex Films. Kistimaat is a 2014 Bangladeshi action film directed by Ashiqur Rahman and produced by Tiger Media Limited and The Abhi Pictures. I'm Taraneh, 15 is a 2002 Iranian film directed by Rasul Sadrameli. The film was released on May 4, 2001.\\
\noindent\textbf{Question}: When was the film Kistimaat released?
\noindent\textbf{Answer}: The film Kistimaat was released in 2014. 
\bigbreak
\noindent Dwaram Venkataswami Naidu and also a lyricist. The film has musical score by S. Hanumantha Rao. Rao died 27 May 1980. Rao married Raja Mani with whom he had three daughters and one son.\\
\noindent\textbf{Question}: Who is the composer of film Baalaraajana Kathe?\\
\noindent\textbf{Answer}: The composer of film Baalaraajana Kathe is S. Hanumantha Rao.
\bigbreak
\noindent Adib Kheir was a leading Syrian nationalist of the 1920s. Filmed on location in the Purcell Mountains in British Columbia, the film was directed by Frank Marshall, written by John Patrick Shanley, and narrated by John Malkovich. Frank Wilton Marshall( born September 13, 1946) is an American film producer and director, often working in collaboration with his wife, Kathleen Kennedy. He received the Irving G. Thalberg award from the Academy of Motion Picture Arts and Sciences in 2018.\\
\noindent\textbf{Question}: Who is the director of film Alive (1993 Film)?\\
\noindent\textbf{Answer}: The director of film Alive is Frank Marshall.
\bigbreak
\noindent[context]\\
\noindent\textbf{Question}: [verifying question]\\
\noindent\textbf{Answer}:

\subsection{Fever}
\subsubsection{Few-shot prompt}
Determine if there is Observation that SUPPORTS or REFUTES a Claim, or if there is NOT ENOUGH INFO.
\bigbreak
\noindent \textbf{Claim}: Reg Watson is a current television producer.\\
\textbf{A}: REFUTES
\bigbreak
\noindent \textbf{Claim}: The Gadsden flag was named by Christopher Gadsden.\\
\textbf{A}: NOT ENOUGH INFO
\bigbreak
\noindent \textbf{Claim}: Black Mirror is about society.\\
\textbf{A}: SUPPORTS
\bigbreak
\noindent \textbf{Claim}: [question]\\
\textbf{A}:
\subsubsection{CoT, CoT-SC prompt}

Determine if there is Observation that SUPPORTS or REFUTES a Claim, or if there is NOT ENOUGH INFO.
\bigbreak
\noindent \textbf{Claim}: Reg Watson is a current television producer.\\
\textbf{A}: First, Reginald James Watson AM was an Australian television producer and screenwriter. Second, Reginald James Watson AM died on 8 October 2019. The answer is REFUTES.
\bigbreak
\noindent \textbf{Claim}: The Gadsden flag was named by Christopher Gadsden.\\
\textbf{A}: First, The Gadsden flag is named after politician Christopher Gadsden. Second, there is no information on who named the Gadsden flag. The answer is NOT ENOUGH INFO.
\bigbreak
\noindent \textbf{Claim}: Black Mirror is about society.\\
\textbf{A}: First, Black Mirror is a British anthology television series. Second, The series uses technology to comment on contemporary social issues. The answer is SUPPORTS.
\bigbreak
\noindent \textbf{Claim}: [question]\\
\textbf{A}:

\subsubsection{Verifying Question Generation prompt}

Write a question that validates the reason for a claim.
\bigbreak
\noindent \textbf{Claim}:  Reg Watson is a current television producer.\\
\textbf{Reason}: Reginald James Watson AM was an Australian television producer and screenwriter.\\
\textbf{Question}: What is Reg Watson's occupation?
\bigbreak
\noindent \textbf{Claim}: The Gadsden flag was named by Christopher Gadsden.\\
\textbf{Reason}: there is no information on who named the Gadsden flag.\\
\textbf{Question}: Who named the Gadsden flag?
\bigbreak
\noindent \textbf{Claim}: [question]\\
\textbf{Reason}: [rationale sentence]\\
\textbf{Question}:

\subsubsection{Verifying Answer Generation (Rationale Editing) prompt}

Reginald James Watson AM (27 August 1926 – 8 October 2019) was an Australian television producer and screenwriter. He was executive producer on Crossroads and created Australian media exports serials such as Prisoner, Neighbours, The Young Doctors and Sons and Daughters.\\
\textbf{Question}: What is Reg Watson's occupation?\\
\textbf{Answer}: Reg Watson was an Australian television producer and screenwriter
\bigbreak
\noindent The flag is named after politician Christopher Gadsden (1724–1805), who designed it in 1775 during the American Revolution.\\
\textbf{Question}: Who named the Gadsden flag?\\
\textbf{Answer}: The Gadsden flag is named after Christopher Gadsden, but there is no information on who named it.
\bigbreak
\noindent [context]\\
\textbf{Question}: [verifying question]\\
\textbf{Answer}:

\section{Human Study}
\label{app:human_study}

\begin{figure}[t!]
    \centering
    \includegraphics[width=0.47\textwidth]{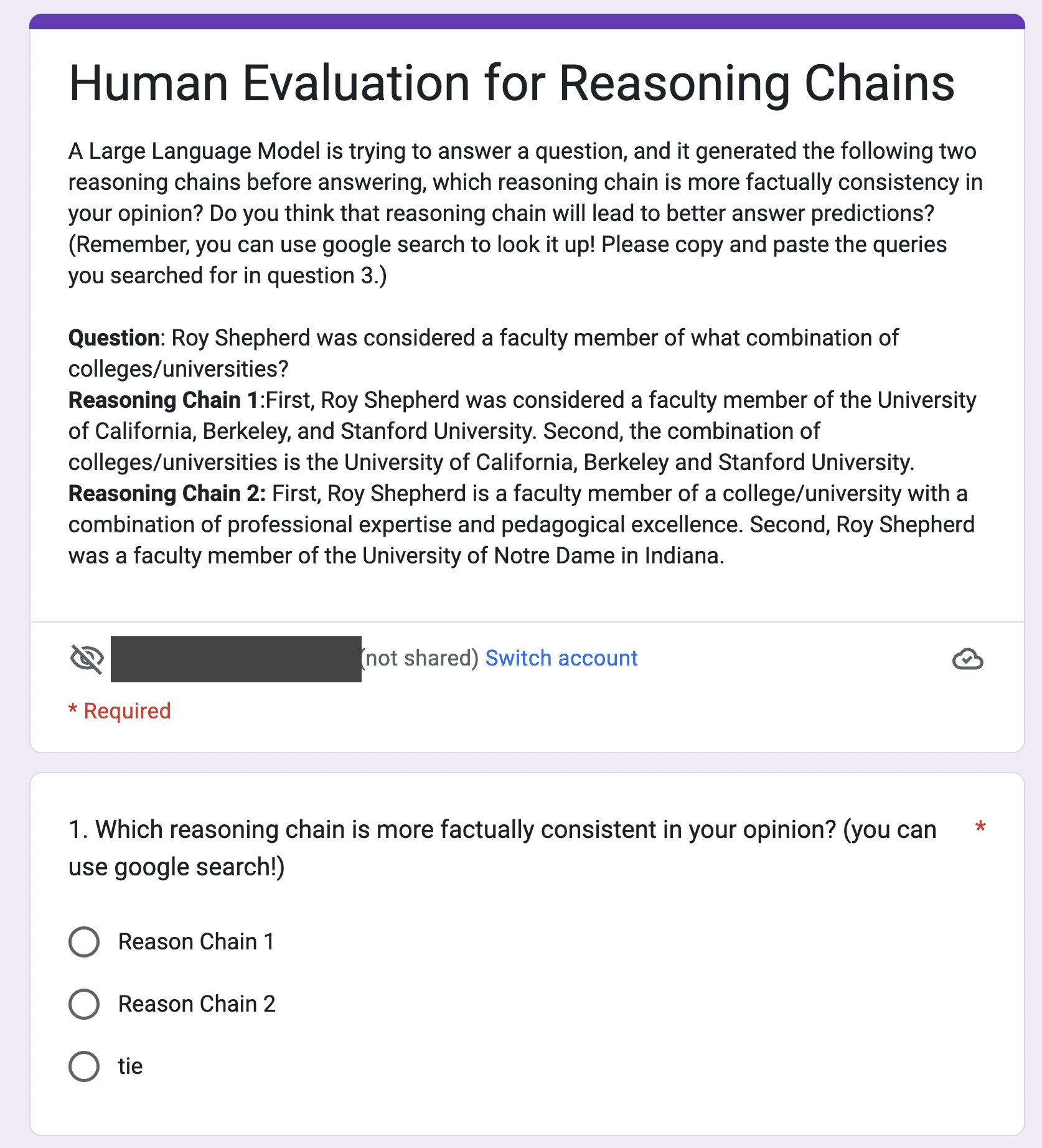}
  \caption{Example Screenshot of Human Evaluation User Interface.}
  \label{fig:human_eval}
\end{figure}

To conduct the human study, we show the instructions in \Cref{fig:human_eval} to two human volunteers. The volunteers are NLP Ph.D. students who are proficient in English. The volunteers understand the use for the data collection and are in consensus. The reasoning chain 1 and 2 are CoTs generated by the CoT-SC baseline and the Verify-and-Edit shown in random order.  On average, each volunteer took 1.25 hours to finish 50 samples.

\section{Qualitative Examples}
\label{app:qualitative_examples}

\begin{table*}[t]
\centering
  \scalebox{0.8}{\begin{tabular}{ll}
    \hline
    Original Question & Suicide's 1977 released album features a song based on what brand's comic character? \\
    True Answer & \textcolor{olive}{Marvel Comics} \\
    CoT-SC & First, Suicide released their 1977 album titled Suicide: Alan Vega \& Martin Rev. \\
    & Second, the album features a song based on the Marvel Comics character Spider-Man.\\
    & The answer is \textcolor{red}{Spider-Man}.\\
    Verifying Questions &  What album was released by Suicide in 1977?\\
    & What comic character is the song on Suicide's 1977 album based on?\\
    Retrieved Facts & Suicide is the debut studio album from the American rock band Suicide. It was released in 1977\\
    & on Red Star Records and produced by Craig Leon and Marty Thau.\\
    & "Ghost Rider" is a song by the protopunk band Suicide appearing on their debut album. The song \\
    & is based on the Marvel Comics character. On Suicide, the noir-movie vibes of engine-revving tunes \\
    & “Ghost Rider” (named after Vega's favorite comic book) and “Rocket U.S.A.” feel scary.\\
    Verify-and-edit & First, Suicide released their debut studio album, titled Suicide, in 1977. \\
    & Second, The song on Suicide's 1977 album is based on the Marvel Comics character Ghost Rider.\\
    & The answer is \textcolor{olive}{Marvel Comics}.\\
    \hline
    Original Question & How many children's books has the writer of the sitcom Maid Marian and her Merry Men written?  \\
    True Answer & \textcolor{olive}{16} \\
    CoT-SC & First, the writer of the sitcom Maid Marian and her Merry Men is Tony Robinson. \\
    & Second, Tony Robinson has written seven children's books.\\
    & The answer is \textcolor{red}{seven}.\\
    Verifying Questions & Who is the writer of the sitcom Maid Marian and her Merry Men?\\
    & How many children's books has Tony Robinson written?\\
    Retrieved Facts & Maid Marian and her Merry Men is a British children's television series created and written by Tony \\
    & Robinson and directed by David Bell. It began in 1989 on BBC1 and ran for four series, with \\
    & the last episode shown in 1994.\\
    & He has published 16 children's books.\\
    Verify-and-Edit & First, The writer of the sitcom Maid Marian and her Merry Men is Tony Robinson.\\
    & Second, Tony Robinson has written 16 children's books.\\
    & The answer is \textcolor{olive}{16}.\\
    \hline
    Original Question & What show other than Hello Ross did Chelsea Handler appear on in January of 2016\\
    True Answer & \textcolor{olive}{Chelsea Does} \\
    CoT-SC & First, Chelsea Handler appeared on the show The Tonight Show Starring Jimmy Fallon in January\\
    & of 2016.\\
    & Second, The Tonight Show Starring Jimmy Fallon is a talk show hosted by Jimmy Fallon.\\
    & The answer is \textcolor{red}{The Tonight Show Starring Jimmy Fallon}.\\
    Verifying Questions & On which show did Chelsea Handler appear in January of 2016?\\
    & What talk show is hosted by Jimmy Fallon?\\
    Retrieved Facts & Chelsea Does is an American streaming television documentary series first released on Netflix on \\
    & January 23, 2016. ... The episodes follow comedian Chelsea Handler as she explores different ...\\
    & The Tonight Show Starring Jimmy Fallon is an American late-night talk show hosted by Jimmy \\
    & Fallon that airs on NBC.\\
    Verify-and-Edit & First, Chelsea Handler appeared on the show Chelsea Does in January of 2016.\\
    & Second, The Tonight Show Starring Jimmy Fallon is a talk show hosted by Jimmy Fallon.\\
    & The answer is \textcolor{olive}{Chelsea Does}.\\
    \hline
\end{tabular}}
\caption{Examples from AdvHotpotQA, facts are retrieved with Google.}
\label{tab:examples}
\end{table*}

In \Cref{tab:examples}, 3 examples from the Adversarial HotpotQA datasets are shown in detail. 

From the first sample, the LLM incorrectly states that the song is ``based on .. Spider-Man.'' However, in the Google retrieved facts, it clearly states that it is based on ``Ghost Rider''. Therefore, the retrieved fact is able to help correct the detail in the rationale. Moreover, although the original rationale also covered the brand name ``Marvel Comics'', the generation goes on with the hero name as an answer, instead of the ``brand'' being asked. Feeding in again also corrects that logical mistake.

In the second example, the LLM makes up a plausible-sounding fact that ``Tony Robinson has written seven children's books''. There is also no indicator on the LLM's confidence level of this claim. Thus, if a user is unfamiliar with this knowledge, it could easily be mistaken as a true fact, which is highly risky. By introducing Google as an assistive tool, we retrieve the sentence ``he has published 16 children's books.'' With this newly retrieved fact in mind, the LLM goes on generating the correct answer. 

The third example is an interesting one. The original CoT already makes mistakes in the first sentence and goes on making continued mistakes in the second sentence as well. This is a type of common mistake in the dataset as well. On correcting them, the Verify-and-Edit framework is able to correct the first claim with the show ``Chelsea Does''. The second claim, however, is verified but irrelevant to the original question anymore. In this case, by feeding in both rationale sentences, the LLM is able to select the relevant fact as an answer, while disregarding the rest. This example shows that the CoT setup used by Verify-and-Edit is important as it allows for models to reason and abstract for a second time, instead of plainly replacing and correcting.

\end{document}